\documentclass[10pt,conference]{IEEEtran}
\IEEEoverridecommandlockouts
\usepackage{cite}
\usepackage{amsmath,amssymb,amsfonts}
\usepackage{algorithmic}
\usepackage{graphicx}
\usepackage{textcomp}
\usepackage{xcolor}
\usepackage{mathtools}
\usepackage[left=0.62in,right=0.62in,top=0.75in,bottom=1.04in]{geometry}
\DeclarePairedDelimiter\ceil{\lceil}{\rceil}
\def\BibTeX{{\rm B\kern-.05em{\sc i\kern-.025em b}\kern-.08em
    T\kern-.1667em\lower.7ex\hbox{E}\kern-.125emX}}
\begin{document}

\title{Modeling Soft-Failure Evolution for Triggering Timely Repair with Low QoT Margins
\thanks{This work has been supported by the European Union’s
Horizon 2020 research and innovation programme under
grant agreement No. 739551 (KIOS CoE - TEAMING) and
from the Republic of Cyprus through the Deputy Ministry of
Research, Innovation and Digital Policy}
}

\author{\IEEEauthorblockN{Sadananda Behera, Tania Panayiotou, Georgios Ellinas}
\IEEEauthorblockA{\textit{KIOS Research and Innovation
   Center of Excellence,} \\ \textit{Department of Electrical and
   Computer Engineering}, \\ \textit{University of Cyprus, Nicosia, Cyprus},  \\
   \{behera.sadananda, panayiotou.tania, gellinas\}@ucy.ac.cy
}
}

\maketitle

\begin{abstract}
In this work, the capabilities of an encoder-decoder learning framework are leveraged to predict soft-failure evolution over a long future horizon. This enables the triggering of timely repair actions with low quality-of-transmission (QoT) margins before a costly hard-failure occurs, ultimately reducing the frequency of repair actions and associated operational expenses. Specifically, it is shown that the proposed scheme is capable of triggering a repair action several days prior to the expected day of a hard-failure, contrary to soft-failure detection schemes utilizing rule-based fixed QoT margins, that may lead either to premature repair actions (i.e., several months before the event of a hard-failure) or to repair actions that are taken too late (i.e., after the hard failure has occurred). Both frameworks are evaluated and compared for a lightpath established in an elastic optical network, where soft-failure evolution can be modeled by analyzing bit-error-rate information monitored at the coherent receivers.  
\end{abstract}


\section{Introduction}
Prevention of hard-failures in optical networks entails the development of effective fault management schemes, ultimately enabling repair actions to take place sufficiently long (but not too long) before the hard-failure is expected to occur~\cite{9714789}. While hard-failures may be caused due to unexpected events (e.g., fiber cuts, physical layer attacks, etc.), their occurrence may be also related to gradual performance degradation (e.g., device ageing, equipment malfunction), that eventually results in the transmission of signals with unacceptable QoT (not detectable at the receiver)~\cite{9422743}. In this work, the focus is on the prevention of the latter type of hard-failures that were shown to be predictable with the use of real data~\cite{Wang:17}, without however excluding the prevention of other types of failures. 

Specifically, the focus is on timely predicting the time that a hard-failure is expected to occur by predicting soft-failure evolution over a long future horizon (e.g., days), enabling repair actions to be taken before soft-failures (e.g., laser drift, filter shift, filter tightening, amplifier malfunction) degenerate to a hard-failure. This is achieved by leveraging the capabilities of an encoder-decoder long short-term memory (ED-LSTM) model~\cite{cho-etal-2014-learning}, designed for sequence-to-sequence problems, capable of mapping an input sequence of past soft-failure observations into an output sequence of future soft-failure estimates.   

Related work on machine learning (ML)-based fault management~\cite{8735762} is by and large categorized into the supervised learning (SL)~\cite{8385774,8869746, 8418780, 8320602, 9483888} and the unsupervised learning (UL)~\cite{9714789} frameworks. In both frameworks, the focus is on the detection, and/or identification, and/or localization of soft-failures, while localization may also concern hard-failures. Regarding soft-failure detection, which is more relevant to this work, in SL this is usually performed according to a binary classifier~\cite{8385774,8869746, 8418780} (i.e., support vector machines (SVMs), neural networks (NNs), random forests (RFs)), trained according to a labeled dataset. This dataset is usually created by inducing soft-failures in an optical network. Hence, the success of this framework is based on a-priori labeling soft-failures, that in practice requires the occurrence of a large number of soft-failures, previously labeled as such (e.g., through fixed QoT margins). To mitigate this problem, UL is applied in~\cite{9714789} to automatically label the dataset to normal and abnormal (i.e., soft-failure) incidents, to subsequently assist the SL approach. 

The main limitation of both approaches is that they do not consider of how these soft-failures evolve over time, possibly leading to inappropriate QoT margins, that on the one hand may lead to premature (i.e., more frequent than necessary) repair actions, and on the other hand to repair actions that are taken too late (i.e., after the hard failure has occurred). In the former case, the a-priori set QoT margins tend to be higher than necessary, eventually increasing operational expenses (OPEX), and in the latter case, the QoT margins tend to underestimate the time that a hard-failure is expected to occur, eventually leading to costly hard-failures. 

Soft-failure evolution is partially considered in~\cite{Wang:17}, by predicting how various observable indicators (e.g., QoT, temperature) are expected to deviate over the next day. This prediction is utilized to decide whether a repair action needs to be proactively taken. The limitation is that the short-term prediction may eventually not allow  a repair action to be timely taken. To ensure the success of this approach, still fixed QoT margins need to be considered over the deviations of the soft-failure indicators. In this work, similar to~\cite{Wang:17}, the focus is on triggering repair actions based on soft-failure evolution. This alleviates the need of a-priori labeling a dataset with normal and abnormal incidents, and most importantly, it alleviates the need of a-priori defining a soft-failure according to a fixed QoT margin. To achieve this, unlike~\cite{Wang:17} that targets short-term predictions, this work targets long-term predictions (i.e., multi-step ahead prediction of soft-failure evolution), ultimately allowing repair decisions to be timely taken with low QoT margins and reduced OPEX. 

 \begin{figure*}[h!]
 \centering
	\includegraphics[scale=0.35]{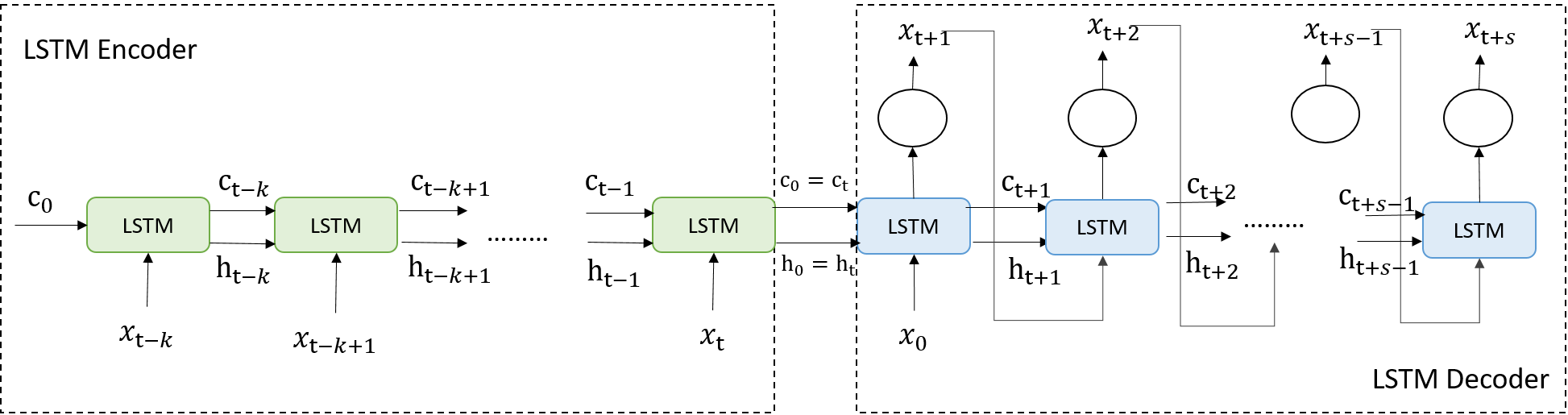} 
	\vspace{-0.1in}
	\caption{A generic ED-LSTM architecture.}
	\label{encoder_decoder}
\end{figure*}
 \begin{figure}[h!]
 \centering
	\includegraphics[scale=0.28]{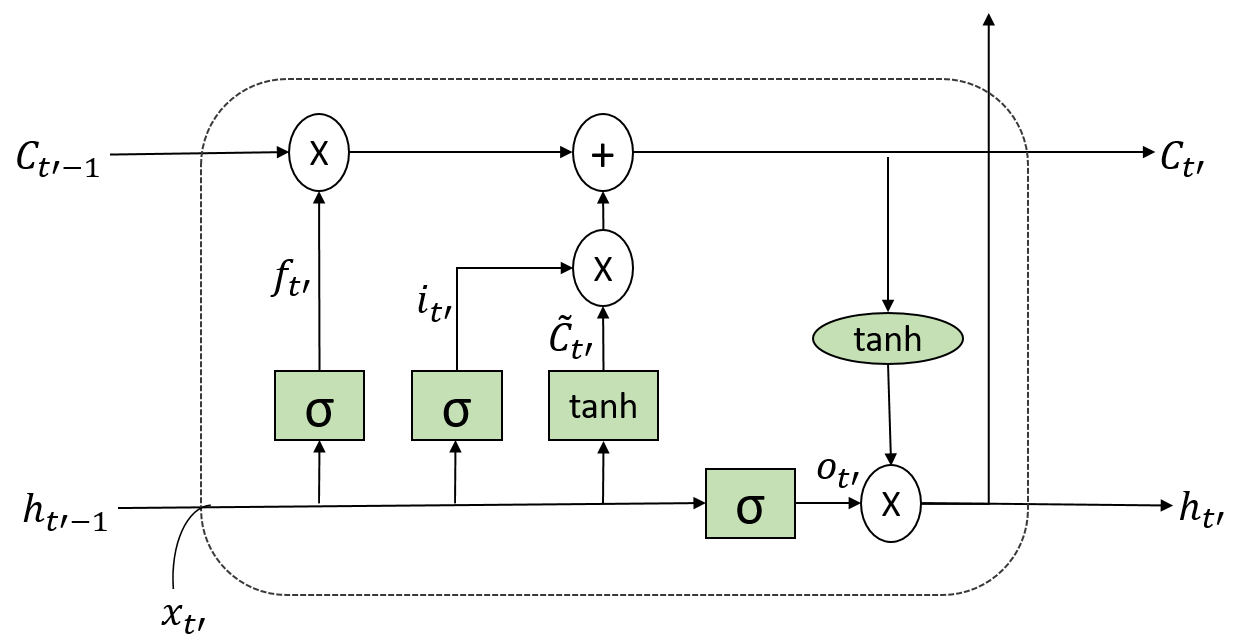} 	
	\vspace{-0.1in}
	\caption{The LSTM cell structure.}
	\label{lstm_cell}
\end{figure}

It  should be noted, that, even though this work focuses on soft-failures that gradually degenerate to hard-failures, long-term soft-failure predictions can also be used for the detection and management of abnormal incidents caused due to other, unexpected events (e.g., attacks); that is, the knowledge of how soft-failures are expected to evolve, allows capturing unexpected network behavior. 

\section{Modeling Soft-Failure Evolution}~\label{model}
The objective of modeling soft-failure evolution with an ED-LSTM can be stated as learning a non-linear function $f(\cdot)$ that, given the past and present QoT observations
\begin{equation*}
x'=[x_{t-\kappa},..,x_{t-1}, x_{t}],
\end{equation*}
accurately predicts the future QoT observations
\begin{equation*} 
x=[x_{t+1}, x_{t+2},..., x_{t+s}],
\end{equation*} 
where $\kappa$ are the past and present QoT observations, $t$ is the present time instant, and $s$ is the number of future prediction steps. In this work, QoT observations in $x'$ are sampled every $\tau$ time units (e.g., every hour) for a total of $k$ past observation windows (e.g., hours), and the aim is to predict the evolution of QoT for the next $s$ windows. Hence, past and future QoT observations are $\tau$ time units apart, with $x'$ and $x$ being sequential in time. 

In general, ED models with recurrent units~\cite{cho-etal-2014-learning}, specifically designed to model sequential data of high dimensionality, are suitable to address the soft-failure evolution problem (i.e., multi-step ahead prediction), especially when the ED is designed with LSTM cells capable of capturing the long-range temporal dependencies of input data. Figure~\ref{encoder_decoder} illustrates a generic ED architecture consisting of LSTM cells, while Fig.~\ref{lstm_cell} illustrates the operation of an LSTM cell. Note that in Fig.~\ref{encoder_decoder}, the LSTM cells are unfolded in time aiming to illustrate their operation over the input and output QoT observations. In practice, however, the encoder-LSTM cell will process an input sequence of $k$ QoT observations sequentially, and the decoder-LSTM cell will predict sequentially the $s$ future QoT estimates.

In general, an LSTM cell, given an input $x_{t'}$ (e.g., the QoT observation upon a time instant $t'$) returns the output hidden state, $h_{t'}$, according to the following recursive equations:
\begin{equation}
{ i}_{t'}=\sigma({ x}_{t'} { U}_i+{ h}_{t'-1}{ W}_i),  
\label{input_gate}
\end{equation}
\begin{equation}
{ f}_{t'}=\sigma({ x}_{t'}{ U}_f+{ h}_{t'-1} { W}_f),
\label{forget_gate}
\end{equation}
\begin{equation}
{ o}_{t'}=\sigma({ x}_{t'} { U}_o+{ h}_{t'-1} { W}_o),  
\label{output_gate}
\end{equation}
\begin{equation}
\tilde{{ c}}_{t'}=\tanh({ x}_{t'} { U}_g+{ h}_{t'-1} { W}_g),
\end{equation}
\begin{equation}
{ c}_{t'}={ f}_{t'} \circ { c}_{t'-1}+{ i}_{t'} \circ \tilde{{ c}}_{t'},
\label{eq_lstm_1}
\end{equation}
\begin{equation}
{ h}_{t'}=\tanh({ c}_{t'})\circ { o}_{t'},
\label{eq_lstm_2}
\end{equation}
where, $\sigma(\cdot)$ is the sigmoid function, $\circ$ is the element-wise product, ${ W}_i \in \mathbb{R}^{u \times u}$, ${ W}_f \in \mathbb{R}^{u \times u}$, ${ W}_o \in \mathbb{R}^{u \times u}$, ${ U}_i \in \mathbb{R}^{d \times u}$, ${ U}_f \in \mathbb{R}^{d \times u}$, and ${ U}_g \in \mathbb{R}^{d \times u}$ are linear transformation matrices (i.e., the unknown parameters to be optimized), $u$ is the number of hidden units, and $d$ is the number of input features. Note that, in this work, where each QoT observation upon any time instant $t'$ is described by a single feature (i.e., the bit-error-rate (BER)), $d$ is equal to $1$. 
Given that, cell state ${ c}_{t'}$ of the LSTM, known as the cell memory, stores a summary of the past input QoT sequence, with the input, forget, and output gate vectors controlling the flow of information in and out of the LSTM cell. Specifically, input gate vector ${ i}_{t'}$ (Eq.~(\ref{input_gate})) is responsible of updating cell state memory, forget gate vector $ {f}_{t'}$ (Eq.~(\ref{forget_gate})) is capable of erasing the cell state memory, and output gate vector ${ o}_{t'}$ (Eq.~(\ref{output_gate})) decides whether to make the output information available or not.

Hence, given the operation of an LSTM cell, the encoder component of the ED model reads each input vector ${x'}$ sequentially, with the cell and hidden states updated according to Eqs.~(\ref{eq_lstm_1}) and~(\ref{eq_lstm_2}), respectively. After reading the end of each input QoT sequence, ${x'}$, the encoder summarizes this input sequence into vectors ${ c}_t$ and ${ h}_t$, subsequently given as inputs to the decoder component, along with a dummy input ${x}_0$ (Fig.~\ref{encoder_decoder}). The purpose of the decoder component is to recursively predict the QoT estimates ${x}_{t+1},.., {x}_{t+s}$ obtained according to ${x}_{t'}=g({ h}_{t'})$ $ \forall t'=t+1, t+2,...,t+s$, where $g(\cdot)$ is the activation function of the output layer. In the soft-failure evolution problem investigated, where the ED-LSTM is trained as a regressor, the output of the decoder is followed by a fully connected dense layer.

In this work, the two components of the ED-LSTM model are trained according to a labeled dataset $D=\{x'^{(i)},x^{(i)}\}_{i=1}^n$, to minimize the mean squared error (MSE) loss function, where $n$ is the number of observed QoT sequences. For training, the Adam optimization algorithm is used~\cite{kingma2014adam}. After model training, the ED-LSTM predicts a future QoT estimate ${x_{t+s'}}$ according to:
\vspace{-0.05in}
\begin{equation}
{x}_{t+s'}= f({x}_{t-k}, \cdots, { x}_{t-1},{x}_t, {x}_{t+1}, { x}_{t+2}, \cdots, {x}_{t+s'-1}),
\end{equation} 
where $s'$ is any future time step and $s'\leq s$.  

Implementation-wise, the dataset can be obtained by a software defined networking (SDN) platform, receiving telemetry data and storing them in a time-series database. The telemetry data are then pre-processed and used for training the ED-LSTM model to provide QoT estimates over the $s$ future time steps, given the past $k$ QoT observations. In the elastic optical network (EON) considered for creating $D$, optical performance monitoring (OPM) information can be obtained from the installed coherent receivers, capable of measuring link impairments in real time (e.g., BER)~\cite{8869746}. For generating $D$, the physical layer model described in Section~\ref{plm} is utilized, in which BER is degraded by gradually inducing soft-failures.  

\section{Physical Layer Model for Dataset Generation}~\label{plm}
\begin{figure}[h]
 \centering
	\includegraphics[scale=0.25]{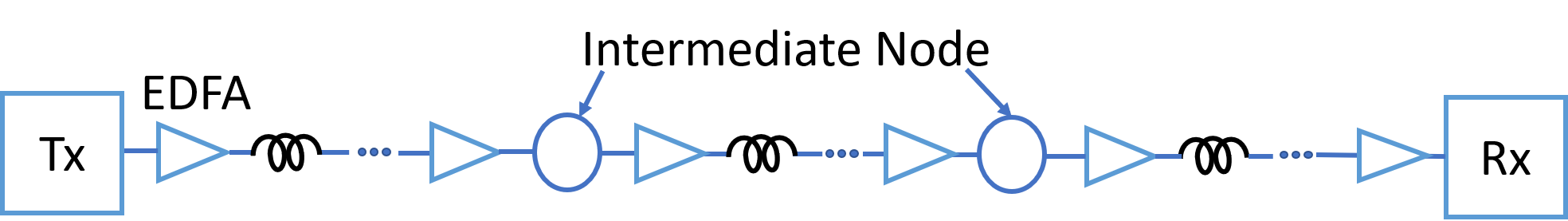} 
	\vspace{-0.1in}
	\caption{Generic simulation setup.}
	\label{testbed}
\end{figure}

 For the generation of BER traces, the generic simulation setup of Fig.~\ref{testbed} is considered. Specifically, in this first attempt to investigate the soft-failure evolution problem, for simplicity, the evolution over a single lightpath is considered, established in an EON with a 6-node mesh topology having nodal degree $3$ and considering fully-loaded links. The considered lightpath is established over a route of $2$ links with distances $400$ and $300$ km, where erbium-doped fiber amplifiers (EDFAs) are spaced $100$ km apart. 
  \vspace{-0.45cm}

\begin{figure}[h]
 \centering
	\includegraphics[width=0.53\textwidth, height=.32\textwidth]{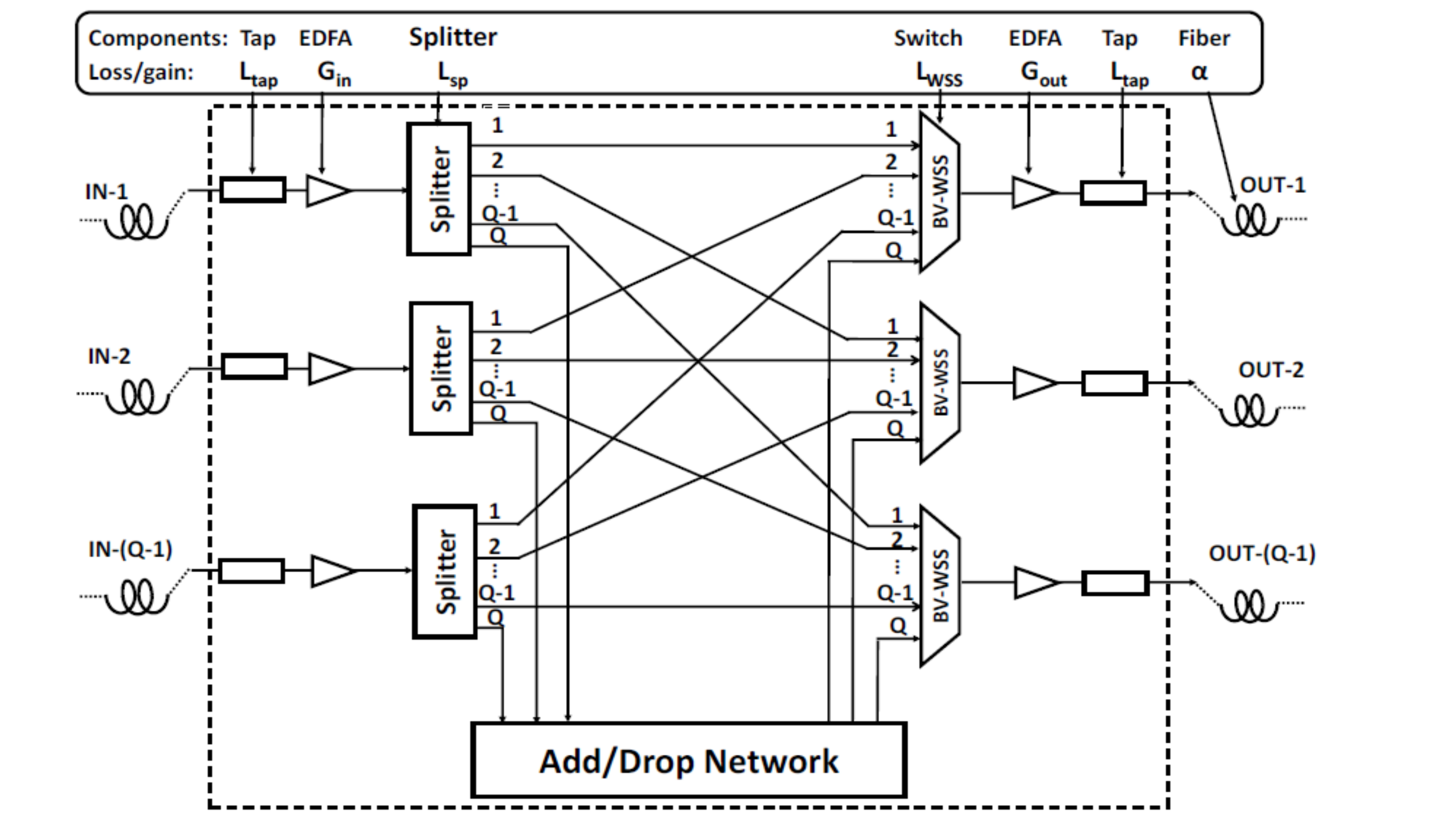}	
	\vspace{-0.2in}
	\caption{Broadcast and select node architecture  with $Q$ inputs/outputs. \cite{behera2019impairment}}
	\label{arch}
\end{figure}

Further, in the EON considered, optical nodes have a typical broadcast and select architecture (see Fig.~\ref{arch}). In this architecture, the in-line EDFA on the input side with gain ($G_{in}$) compensates fiber attenuation ($\alpha$) and tap losses ($L_{tap}$). The booster EDFA with gain ($G_{out}$) compensates splitter loss ($L_{sp}$) and switch loss ($L_{WSS}$). Therefore, $G_{in}=\alpha \times l+L_{tap}$ and $G_{out}(z) \geq 3 \ceil{log_{2}Q(z)}+L_{WSS} \hspace{0.2cm} \text{dB}$, where $l$ is the link length, and $Q(z)$ is the fiber input/output port  at node $z$. Moreover, the following parameters are defined for the calculation of the SNR in route $r$ of a lightpath from source $s$ to destination $d$: ${H}_{{r}}^{{s,d}}$ (no. of hops), ${M}_{{r}}^{{s,d}}$ (no. of EDFAs), $P_{r}^{s,d}$ (received power), and $N_{r}^{s,d}$ (ASE noise power). The received signal power (in dBm) at the $i^\text{th}$ time instance for a lightpath from node $s$ to node $d$ is given as,
\begin{align}
	\label{pr}
	 P_{r}^{s,d}(i)=&P_t-\underbrace{(M_{r}^{s,d}\times \alpha l)-L_{WSS}-L_{tap}}_\text{Total loss}\nonumber  \\
	 &+\underbrace{G_{in}(i)}_\text{\parbox{1cm}{\centering Gain of\\[-4pt] degrading EDFA}}
	+\underbrace{\bigg((M_{r}^{s,d}-1)\times G_{in}(1)\bigg)}_\text{Gain of non-degrading EDFAs} \nonumber  \\
	&+\underbrace{\sum_{j=1}^{{H}_{{r}}^{{s,d}}-1}G_{out}(j)}_\text{Gain of booster EDFA}
\end{align}

Similarly, the accumulated ASE noise for a lightpath from node $s$ to node $d$ over route $r$ is given as,
\begin{align}
	\label{np}
	N_{r}^{s,d}(i)=&\gamma \bigg[ n_{sp_{in}}\bigg((M_{r}^{sd}-1)(G_{in}(1)-1)\nonumber\\
	&\hspace{-1cm}+(G_{in}(i)-1)\bigg)	+ n_{sp_{out}}\sum_{j=1}^{H_{r}^{s,d}-1}(G_{out}(j)-1) \bigg]	
\end{align}
\text{where,} $G_{in}(1)$ is the EDFA gain before degradation and $\gamma=2hf_{c}B_{e}$ (see Table \ref{sim param} for each term). Then, the SNR can be calculated using Eqs. \eqref{pr} and \eqref{np}. Note that 4-QAM modulation is also assumed and that, given the SNR, the BER for this case is given as,
\begin{equation}
BER_{4-QAM}=\frac{1}{2}erfc\left(\sqrt{\frac{SNR}{2}}\right).
\label{4qam}
\end{equation}
 \vspace{-0.25cm}
\begin{table}[h]
	\centering
	\caption{PHYSICAL LAYER PARAMETERS}
	 \vspace{-0.05cm}
	\label{sim param}
	\begin{tabular}{|l|l|}
		\hline
		{\bf Parameter}                                  & {\bf Value}     \\ \hline
		Transmit power ($P_t$)                        & -17 dBm   \\ \hline
		Operating frequency ($f_c$)                      &193.1 THz   \\ \hline
		Spontaneous emission            & 3        \\ 
		factor for in-line EDFA ($n_{sp_{in}}$) & \\ \hline
		Spontaneous emission            & 2        \\ 
		factor for booster EDFA ($n_{sp_{out}}$) & \\ \hline
		Fiber attenuation ($\alpha$)                       & 0.2 dB/km\\ \hline
		WSS loss $(L_{WSS})$ &  2 dB \\ \hline		
		Tap loss ($L_{tap}$) & 1 dB \\ \hline
		EDFA spacing                            & 100 km   \\ \hline		
		Output EDFA gain ($G_{out}$)                             & 8 dB  \\ \hline  
		Electrical bandwidth ($B_e$)      & 7 GHz      \\ \hline
		Planck's constant (h)  & $6.62 \times 10^{-34}$ J.s \\ \hline
		\end{tabular}
\end{table}

To induce soft-failures, one in-line EDFA is selected, and its gain is gradually degraded, while keeping all other EDFA gains (both in-line and post EDFAs) constant. It is assumed that all EDFAs are of the same type, therefore, selection of any EDFA can represent the proof of concept for the application of the presented model. The physical layer parameters used can be found in Table \ref{sim param}, while the gradual degradation of the EDFA's gain is described in Section~\ref{dataset_generation}. Note that, for the simulation setup utilized, the presence of one type of soft-failure is assumed; that is, EDFA gain degradation. However, this assumption does not affect the scope of this work, since soft-failure evolution is modeled according to the accumulated information at the receiver. The combination of several soft-failures can be considered as well, with their impact similarly captured through OPM and associated indicators (i.e., features) for ED-LSTM model training. 
  
\section{Dataset Generation and Pre-processing}~\label{dataset_generation}
\begin{figure}[h!]
 \centering
	\includegraphics[scale=0.45]{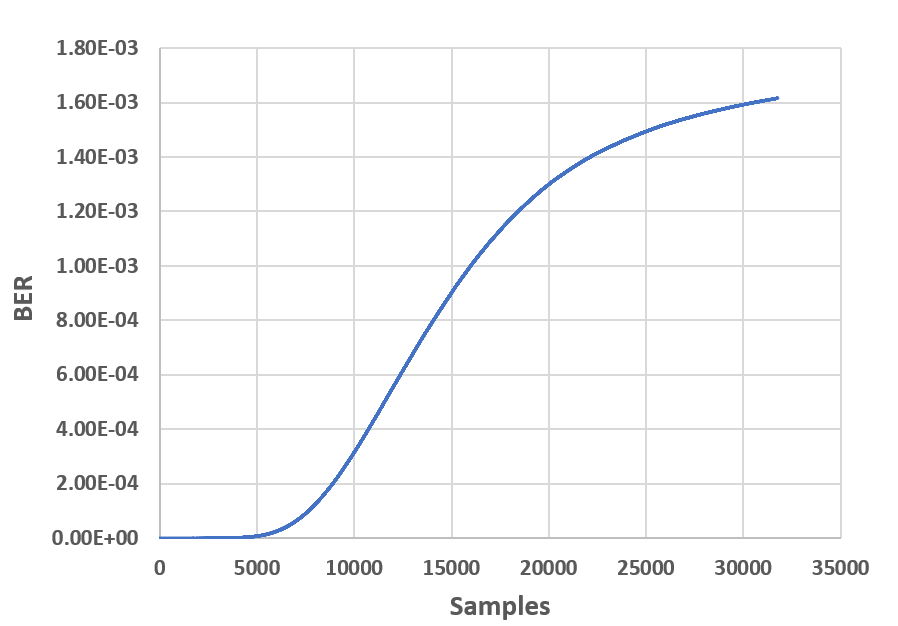} 
	\vspace{-0.1in}
	\caption{Example of BER evolution over time.}
	\label{Dataset}
\end{figure}

To obtain dataset $D$, the lifetime of an EDFA is modeled according to the Weibull distribution $W(\lambda, \beta)$, capable of describing the aging effects of the network
components~\cite{1354653}, and EDFA gain traces  with a fixed degradation rate $\delta$ are generated. Specifically, a sequence of time stamps is sampled from distribution $W$, where EDFA gain degradation occurs with rate $\delta$. This work considers an initial EDFA gain $G_{in}(1)=22$ dB, degradation rate $\delta=10^{-6}$, and scale and shape parameters of the Weibull distribution $\lambda=595.75$ and $\beta=1.05$, respectively. An example of BER evolution over time is shown in Fig.~\ref{Dataset}, where samples are sequential in time and follow the Weibull distribution. According to Fig.~\ref{Dataset}, BER degradation is slower at the beginning (i.e., when the device is new) and the rate of degradation increases over time (i.e., with ageing).   

Specifically, for generating BER samples, for each degradation step, the SNR is evaluated and is mapped into the corresponding QoT (in terms of BER - Eq.~\eqref{4qam}). In total, $1$ million sequential-in-time samples are generated to create an initial dataset $D'$. To create the training/test dataset $D$, a time scale of $\tau =90$ minutes is considered, according to which dataset $D'$ is partitioned into windows consisting of sequential-in-time samples that span a period of $1.5$ hours. Hence, $D$ is created by sampling sequentially from each window in $D'$ the last sample, in order to create sequences of QoT observations that are $1.5$ hours apart. Specifically, each sequence $[{x}^{(i)}_{t-k}, \cdots, { x}^{(i)}_{t-1},{x}^{(i)}_t, {x}^{(i)}_{t+1}, { x}^{(i)}_{t+2}, \cdots, {x}^{(i)}_{t+s}]$ is created with $k=50$ and $s=70$, following the sliding window approach (i.e., sequence $i$ shares $119$ common samples with sequence $i-1$).

Overall, the total number of sequences considered in $D$ is $n=6081$, spanning an one year period. Hence, the objective of the ED-LSTM model is to find a model, that, given as input $50$ past and the present QoT observations, accurately predicts the QoT values $70$ steps ahead; that is, the aim is to predict the soft-failure evolution $4$ days ahead of time, ultimately allowing a network operator to timely initiate repair actions (i.e., identify and localize the failure and dispatch technicians to the site) if the predicted QoT is close to a threshold violation (i.e., close to a hard-failure). Note that, in practice, and according to the problem setting described, for the creation of dataset $D$ the QoT value  only needs to be monitored at the coherent receivers every $1.5$ hours. Depending, however, on the requirements of the network operator on repairing a soft-failure (i.e., time requirements), $h$, $s$, and $\tau$ parameters can be set accordingly to allow for timely repair actions.  

Finally, it is important to note that dataset $D$ does not include all the sequences that can be extracted from the initial dataset $D'$; rather, only a subset of these sequences is considered that spans only a period of one year. This is necessary for the ED-LSTM model to converge to an accurate model, since sequences in $D'$ are non-stationary and may significantly vary over time (this is obvious from Fig.~\ref{Dataset}). This is mainly due to the consideration of the Weibull distribution, which mimics the realistic lifetime of devices (i.e., degradation time steps are more frequent as time evolves). Hence, to overcome this limitation, and accurately model soft-failure evolution over non-stationary data, in practice, the ED-LSTM model must be appropriately re-trained according to an updated dataset that is shifted in time (i.e., includes the most recent information). However, this does not hinder the applicability of the proposed approach, since training is performed off-line within less than an hour, while the ED-LSTM model becomes obsolete after a period of several months (i.e., needs re-training after several months). The latter outcomes are further discussed in the next section, that focuses on model training and evaluation.      

\section{Model Training and Evaluation}
For ED-LSTM model training and testing, $D$ is split in such a way that $90\%$ of the sequences is used for training, $20\%$ of which is used for validation, and $10\%$ is used for testing. Note that training and test datasets are sequential in time, with the training dataset including the first $5472$ sequences of $D$, and the test dataset including the last $609$. Furthermore, the training dataset includes only sequences with adequate QoT (i.e., a hard-failure is not present in the training dataset). Regarding the ED-LSTM architecture, both the encoder and decoder components are designed with one hidden layer, consisting of $u=30$ hidden units. The output of the LSTM decoder is wrapped by a time-distributed dense layer with $20$ units. For training, the learning rate is set to $10^{-5}$, the batch size to $16$, and the number of epochs to $500$. The ED-LSTM requires $50$ minutes of training and validation  time in the computation system utilized, with Intel Core i5-6500 CPU @3.2 GHz and 8 GB RAM. Note that the ED-LSTM model was tested according to various other hyperparameters (i.e., number of hidden units, learning rate, etc.) and datasets spanning a larger period of time, and the aforementioned configuration resulted in the most accurate model. 

\begin{figure}[h!]
 \centering
	\includegraphics[scale=0.45]{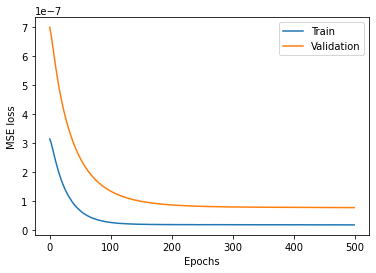} 	
	\vspace{-0.1in}
	\caption{ED-LSTM training/validation MSE loss versus epochs.}
	\label{train_validn_10}
\end{figure}

Figure~\ref{train_validn_10} illustrates ED-LSTM training/validation evolution over the MSE loss. It is observed that both training and validation converge to an MSE loss that is close to zero. A small deviation between the training and validation loss is, however, observed, mainly due the non-stationary nature of the samples in the dataset (refer to  Fig.~\ref{Dataset}). This deviation is, however, negligible and it does not affect the accuracy of the model over the test dataset. Specifically, model accuracy over the test dataset is $1.26\times 10^{-7}$, which is close to the validation accuracy after all $500$ epochs (i.e., close to zero).

\begin{figure}[h!]
 \centering
	\includegraphics[scale=0.45]{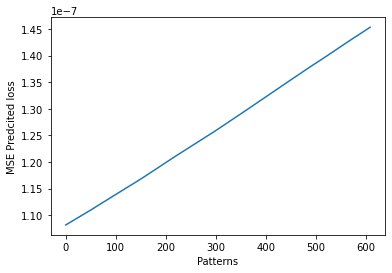} 	
	\vspace{-0.1in}
	\caption{MSE predicted loss with respect to future patterns.}
	\label{MSE_10}
\end{figure}

To gain a better insight on the accuracy of the model over the test dataset, Fig.~\ref{MSE_10} illustrates the MSE predicted loss per pattern over each unseen future sequence (i.e., not used during training). As expected, this prediction loss is lower for the future sequences that are closer in time to the sequences used for training the model. However, as the prediction horizon increases, the prediction loss increases as well, since BER degradation is non-stationary over time. Nevertheless, for the prediction horizon considered in the test dataset, prediction loss is still adequate, rendering the proposed model sufficient for decision-making as it concerns repair decisions. Importantly, however, observation of the loss deviations over the predictions can be used for triggering model re-training, by shifting (in time) the training dataset to subsequently increase model accuracy. For model re-training, transfer learning techniques can be exploited to reduce both the number of samples and the time required for model convergence~\cite{9253990}.

\section{Comparative Analysis}
In this section, a comparative analysis is provided to demonstrate the importance of considering soft-failure evolution for triggering repair actions. Specifically, the proposed approach is compared with the common approach followed in schemes that do not consider soft-failure evolution~\cite{8385774,8869746, 8418780, 8320602, 9483888,9714789} for proactively triggering a repair action. In these schemes, a soft-failure is commonly identified (i.e., classified) according to QoT deviations that are caused over a range of induced degradations in one or various devices along the optical connections. It is shown, however, that such QoT deviations may lead to QoT margins that either
overestimate or underestimate the time that a repair action needs to be taken (i.e., how critical is the soft-failure). 

Specifically, to provide an analysis that is directly comparable with the type of soft-failures considered in this work (i.e., EDFA gain degradation), gain degradation is randomly induced in the system, that ranges between $5-10$ dB (i.e., a typical range of values, previously considered in related works \cite{9483888,8735762}). The corresponding QoT degradation is then obtained and, subsequently, the times that this QoT degradation triggered a repair action are observed in the initial dataset $D'$. Additionally, the margin between the soft- and hard-failure QoT values is evaluated (as a percentage). Note that, for the purposes of this comparative analysis, the hard-failure is set according to a QoT threshold that is equal to $10^{-3}$; that is, once a higher BER is observed then this is the time that the hard-failure occurs. Note that, in practice, a hard-failure can be defined to include a small QoT margin in order to consider, for example, uncertainty over the future estimates (e.g., through quantile or Monte Carlo inference techniques~\cite{MARYAM2022108992, Maryam22}). 
\begin{table}[h!]
\centering
\renewcommand{\arraystretch}{1.5}
\caption{Comparative Results}
 \vspace{-0.05cm}
\label{comp}
\begin{tabular}{|p{2cm}|p{2.5cm}|p{2.5cm}|}
\hline

 \multicolumn{3}{|c|}{\bf Fixed Soft-Failures} \\ 
Gain Reduction   &  Repair Action &  QoT Margin \\ \hline
5 dB                 & 65 days ahead             & 32\%                                                                              \\ \hline
7 dB                  & 42 days ahead              & 17.11\%                                                                           \\ \hline
10 dB                    & Hard-failure \newline occurred & Hard-failure \newline occurred                                                             \\ \hline \hline
 \multicolumn{3}{|c|}{\bf Predicting Soft-Failure Evolution} \\
Gain Reduction   &  Repair Action &  QoT Margin \\ \hline 
9.06 dB  & 4 days ahead           & 5.32\%                                                                            \\ \hline
\end{tabular}
\end{table}   

A summary of these comparative results is provided in Table~\ref{comp}. Specifically, Table~\ref{comp} illustrates the results of indicative gain reductions (i.e., $5$, $7$, $10$ dB) that lead to fixed soft-failure thresholds and the results obtained from the ED-LSTM model that predicts the soft-failure evolution. According to these results, the scenarios with $5$ and $7$ dB gain reductions overestimate the time that a repair action is taken (i.e., $65$ days and $42$ days before the hard-failure is expected to occur). On the contrary, for the scenario with a $10$ dB gain reduction, action is taken after the occurrence of the hard-failure, which means that the consideration of such a fixed soft-failure underestimates the time that an action needs to be taken. On the contrary,  the proposed scheme, which is trained to predict the soft-failure evolution $4$ days ahead of time, is capable of timely triggering a repair action, while the appropriate soft-failure threshold does not need to be known a-priori. However, for comparative purposes, the corresponding gain reduction was obtained (i.e., $9.06$ dB), which leads to a low QoT margin that is just $5.32\%$ below the nominal pre-FEC BER requirement for the lightpaths. Comparatively, fixed QoT soft-failures correspond to a QoT margin that is $32\%$ ($17.11\%$) below the nominal pre-FEC BER requirement when the scenario with $5$ ($7$) dB reduction is considered. Overall, these results indicate that by predicting soft-failure evolution over an adequately large future horizon allows a network operator to timely trigger repair actions, and importantly, it does so with low QoT margins. In general, low QoT margins reduce the frequency that repair actions are taken, consequently reducing OPEX. 

\section{Conclusions}
This work investigates an ML-aided framework based on ED-LSTM models, for predicting soft-failure evolution over a long future horizon. It is shown that the ED-LSTM model enables a network operator to accurately predict the time that a hard-failure is expected to occur, ultimately triggering a timely repair action (i.e., 4 days prior to the hard-failure in this case) and with a low QoT margin. The performance of the proposed model is compared with traditional rule-based soft-failure schemes that, unlike the proposed approach, are shown to overestimate or underestimate the time that a repair action needs to be taken; that is, depending on the QoT margin considered, a repair action is taken too early (i.e., up to 65 days prior to the hard-failure) or too late (i.e., after the  occurrence of the hard-failure). Importantly, the proposed method is shown to be capable of triggering a repair action according to a low QoT margin that deviates only $5.32\%$ from the hard-failure QoT threshold, while QoT margins of rule-based schemes are shown to deviate up to $32\%$. This improvement, ultimately reduces the frequency of repair actions and in turn the network's OPEX.

The proposed soft-failure evolution framework constitutes the basis for many interesting future directions, including the consideration of other types of soft-failures and the presence of several lightpaths and their inter-dependencies on more complex EON topologies~\cite{9623420}. Further, the knowledge of how soft-failures evolve over time can be used for soft-failure identification and localization, while appropriate mechanisms can be developed to identify unexpected QoT deviations (e.g., attacks). Considering the uncertainty over future soft-failure estimates also constitutes an interesting future direction.    

\bibliographystyle{IEEEtran}
\bibliography{IEEEabrv,references_v2}

\end{document}